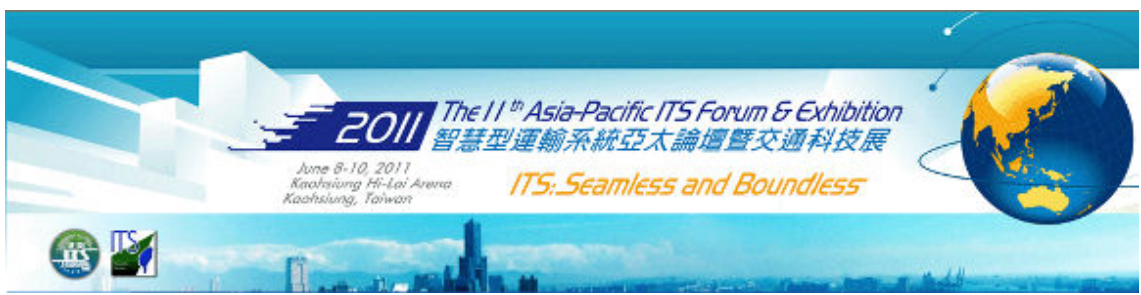

# Image-based Vehicle Classification System

**Jun Yee Ng[1], Yong Haur Tay[2]**

*Universiti Tunku Abdul Rahman, Kuala Lumpur, Malaysia*
[1]*junyeeng@yahoo.com*, [2]*tayyh@utar.edu.my*

**Abstract**

Electronic toll collection (ETC) system has been a common trend used for toll collection on toll road nowadays. The implementation of electronic toll collection allows vehicles to travel at low or full speed during the toll payment, which help to avoid the traffic delay at toll road. One of the major components of an electronic toll collection is the automatic vehicle detection and classification (AVDC) system which is important to classify the vehicle so that the toll is charged according to the vehicle classes.

Vision-based vehicle classification system is one type of vehicle classification system which adopt camera as the input sensing device for the system. This type of system has advantage over the rest for it is cost efficient as low cost camera is used. The implementation of vision-based vehicle classification system requires lower initial investment cost and very suitable for the toll collection trend migration in Malaysia from single ETC system to full-scale multi-lane free flow (MLFF).

This project includes the development of an image-based vehicle classification system as an effort to seek for a robust vision-based vehicle classification system. The techniques used in the system include scale-invariant feature transform (SIFT) technique, Canny's edge detector, K-means clustering as well as Euclidean distance matching. In this project, a unique way to image description as matching medium is proposed. This distinctiveness of method is analogous to the human DNA concept which is highly unique. The system is evaluated on open datasets and return promising results.

Key words: Electronic toll collection (ETC), vehicle detection and classification (AVDC), multi-lane free flow (MLFF), Scale-invariant feature transform (SIFT)

## 1. Introduction

Electronic toll collection (ETC) has been a common trend in road pricing system all over the world. The technology aims to eliminate the delay on toll roads by collecting tolls electronically. Until today, many countries are implementing the road pricing concept with different approaches. Most of the electronic toll collection adopted radio frequency identification technology. In some urban setting, the automated gates used in the electronic toll lanes allows vehicle to travel at a limited speed. More sophisticated setting allows vehicles to travel at full speed. This directly eliminates the delay on toll roads which is a may lead to traffic jam. Among the advantages of the implementation of electronic toll collection is the independence of man involvement in toll collection system, high efficiency, the implementation of criminal vehicle tracking into the traffic management system, and serves as part of the components in intelligence transport system.

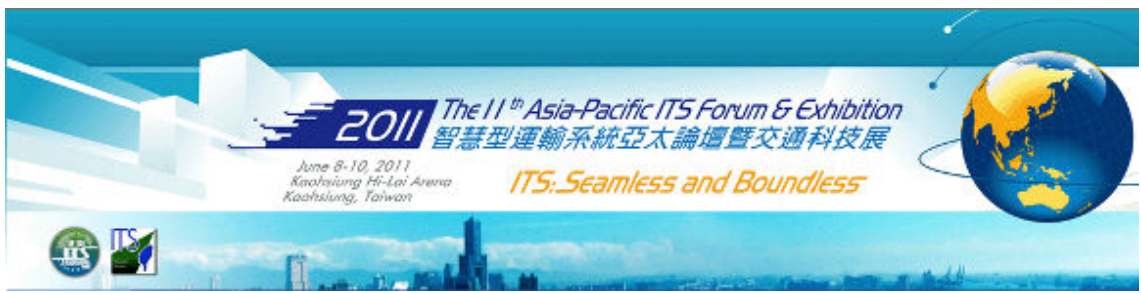

Electronic toll collection scheme consist of four major components: automated vehicle identification, automated vehicle classification, transaction processing, and violation enforcement. This paper focuses on the automatic vehicle classification system. Automatic vehicle classification system classifies a vehicle detected according to the types of the vehicle. It has been an active research for the past decade.

To date, many vehicle classification methods have been proposed. These methods vary mostly by the type of sensors used. These sensors include magnetic sensor, laser sensor, camera, multiple calibrated cameras etc. Vehicle detection and classification (AVDC) system with magnetic and laser sensor has better accuracy in the sense that the 3 dimensional profile of the vehicle can retrieved from the sensor. However, these sensors are costly and the system requires a large amount of start-up expenses. Vision-based AVDC system uses single or various cameras to capture the two-dimensional image or image sequences of vehicles as the input for the system. Vision-based AVDC system requires lower start-up expenses, but have relatively high error rate due to environmental factor such as shadow, night view, occlusion, and others. Most of the existed vision-based AVDC system adopted the shape, length and height of the vehicles as the features for the classifying task while more robust approach uses of keypoints detection algorithm.

In Malaysia, the key challenges of MLFF implementation is migrating from the current ETC to a full-scale multi-lane free flow which would require large amount of initiation cost. This initiation cost is related to the construction of transponder as well as the installation tag in every vehicle. To tackle this, automatic vehicle classification system can be used along with the currently used RFID tag pricing system in the country.

In this project, an image-based vehicle classification system is developed as an effort to find out the solution to the challenges of MLFF implementation in Malaysia. This project also aims to find out a robust framework for vehicle classification system. The system developed in this project adopted Scale Invariant Feature Transform, K-means clustering, as well as Euclidean distance matching as part of the components for the system. The system developed is tested with an open source dataset provided in a related study and returns promising result.

2. Related Research

Over the decade, vehicle classification system has been an active research. Many methods have been proposed using different approach and sensors. These methods can be categorized by the sensors used in the classification system, namely laser sensors, magnetic sensors, and vision sensors.

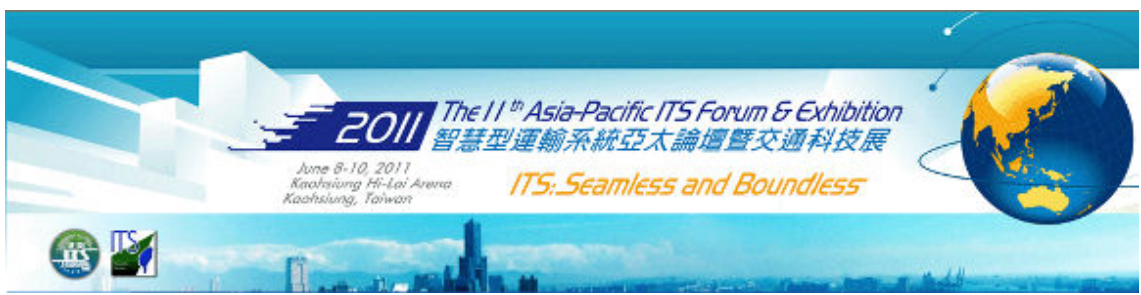

Laser sensors have the highest reliability because it allows the three dimensional profile of the vehicle to be retrieved as the information for vehicle classification task [1]. An operation conducted by Tropartz et al [13] in 1999 shows that laser sensor has very high reliability as well as accuracy. The operation involves an overhead laser sensors evaluated under difficult traffic condition. The evaluation of 2.3 million vehicle classifications for vehicle passages at 8 commercial toll plazas yielded a mean classification accuracy of 98.5 %. Hussain and Moussa, (2005) [4] states that the uses of laser range sensors offer the promise of sensors that are less sensitive to deteriorated environmental conditions such as rain and fog. However, laser sensors are very expensive compared to vision-based sensor. It would require an extremely large amount of initiation cost if such vehicle classification system was to be implemented.

Inductive sensor or magnetic sensor can be use to extracts dominant low-complexity features including vehicle count, speed, length, Hill-pattern peaks, and normalized energy [7]. An evaluation conducts in Kaewkamnerd et al in 2009 [7] shows that the result when the classification is based on sizes return an accuracy of (95%). But when classification is done on vehicles of similar sizes (sedan car, van pickup truck), the result is as low as 70% - 80%.Similiar evaluation was performed in Kaewkamnerd et al (2010) [6] in which the result gained is slightly better, 97% and 77% for each cases as stated above.

Vision-based vehicle classification system uses low cost camera to capture image or image sequences of the scene of a vehicle. Then, the classification system retrieves the information from the image based on the algorithm used. Therefore, the performance of a vision-based vehicle classification system relies very much on the algorithm instead of the sensor. Vision based sensor can either be image-based or video based. Video-based camera has advantage over image-based sensor because moving object can be separated from the static background reasonably well by background modelling and substraction [10]. This is termed as segmentation whereby the vehicle is segmented from the background for further processing. Segmentation can be limited by environmental factors such as shadow, reflection, fog and rain. Shadow elimination can be included as part of the algorithm to solve the shadow problem. In many conventional vision-based vehicle classification systems, the features of the vehicle used for the classification task are the height, length and width of the vehicle [2, 3, 12]. More sophisticated algorithm adopted the keypoints detection on the vehicles as the features for the classification task [10]. Keypoint as the features for classification has advantage over the conventional dimensional profile due to the high reliability of the scale invariant keypoint detection algorithms.

Scale-Invariant Feature Transforms (SIFT) by David Lowe in 1999 [8] is a widely used keypoint detection algorithm. The method is notable for the reason that the features used are invariant to image scaling, translation, and rotation, and partially invariant to illumination changes and affine or 3D projection [8, 9]. In Ma and Grimson (2005) [10], the keypoint detection algorithm is adopted and modified to fit in vehicle classification task and return a good result.

3. **System Overview**

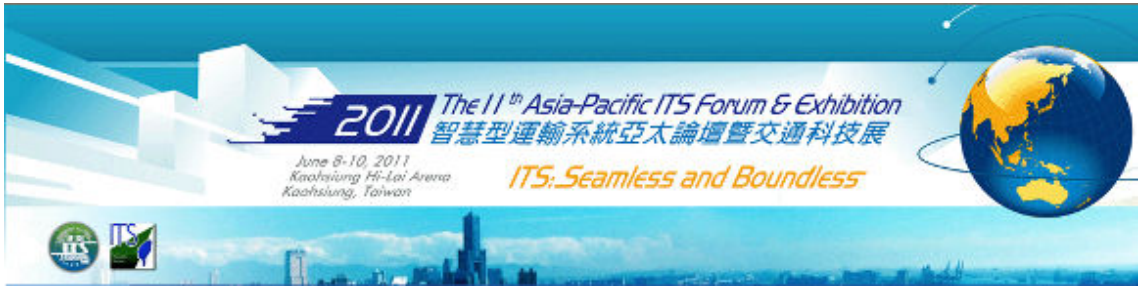

This section discuss about the overview of the vehicle classification system developed in this project. Two vehicle classification systems are developed to carry out two vehicle classification tasks: Inter-class vehicle classification (Car/Van classification) and Intra-class vehicle classification (Sedan/Taxi classification) as shown in Figure 3.1.

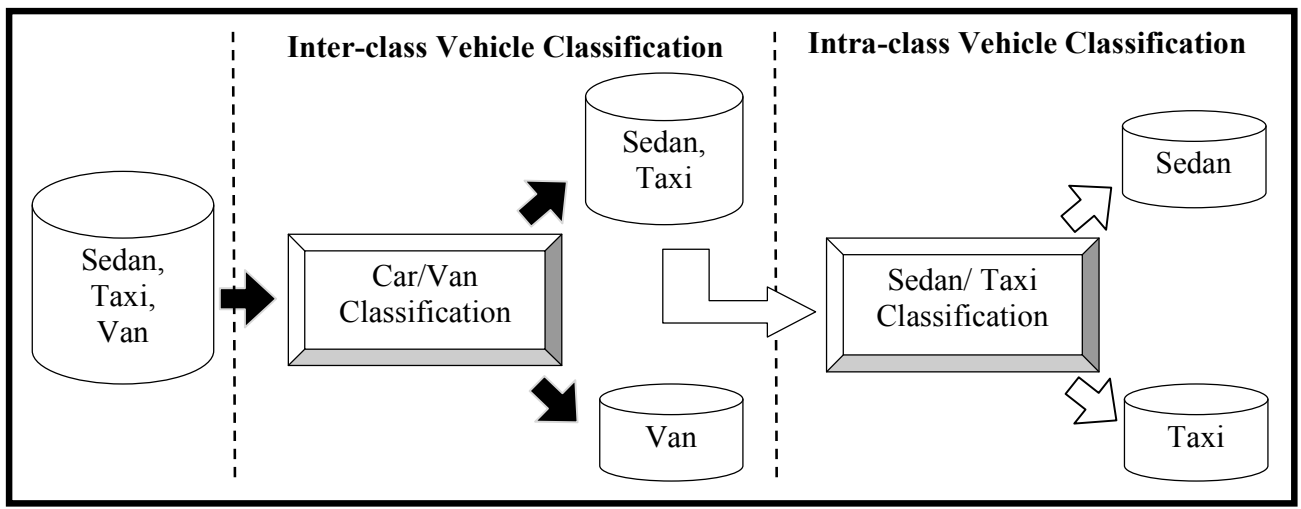

Figure 3.1: The two classification task: Inter-class classification (car/van classification) and intra-class classification (taxi/sedan car classification).

The framework of each system is divided into two parts: training scheme and matching scheme. In the training scheme, the system is trained with training image to be intelligent. The output from the training serves as the input for the matching scheme. The training scheme is meant for offline performance, which means it only occurs one time in the system prior to any real time task. In matching scheme, the system takes in a query input. Based on the output obtained from the training scheme, the query input is processed to find out the vehicle type. This is done on instant real time basis after the system training.

a. **Inter-class Vehicle Classification (Car/Van classification)**

Inter-class classification (car/van classification) is the easier task to perform between the two tasks. This is due to the distinctive features that can be found between the two classes of vehicles.

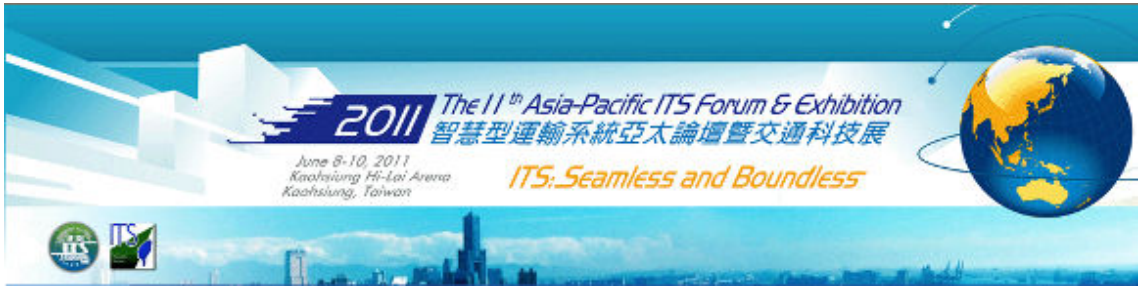

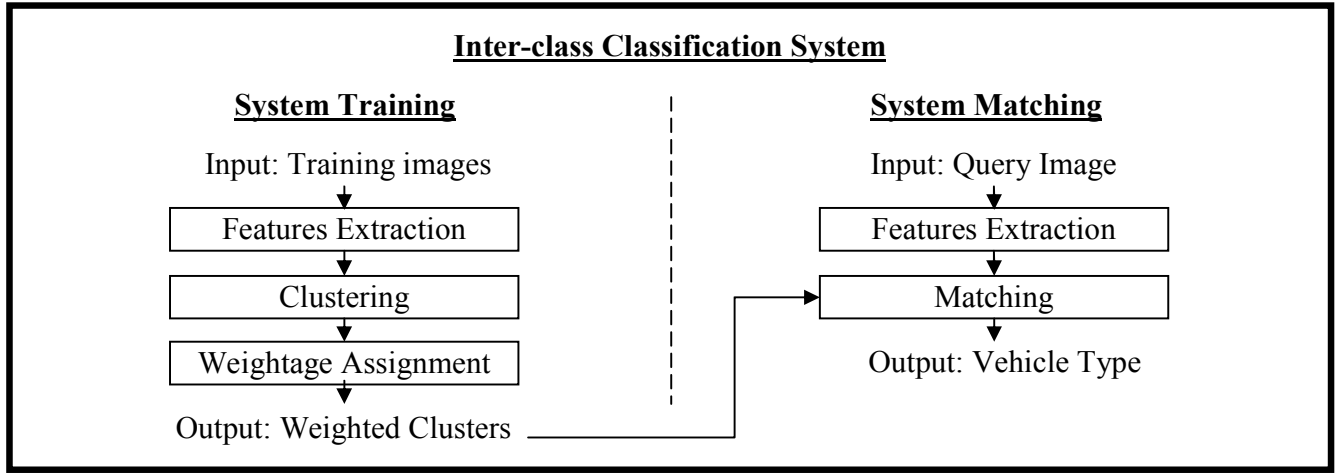

Figure 3.2: The training and matching scheme for the inter-class vehicle classification system

As shown in Figure 3.2 above, the system is feed with selected training images during the system training. All the training images are processed and the features of the vehicle are extracted into the system. The features extracted can be redundant; therefore, the clustering takes place right after the feature extraction to cluster the output. After getting all the features in clusters from the training image, the system then assigns weightage onto each cluster. These weighted clusters by the end of the training serves as the input for the system for matching purpose.

In system matching scheme, the system is feed with a query image in which the vehicle type contained is to be determined. The input undergoes feature extraction and all the features extracted from the image are matched with the weighted clusters from the training. From the matching, a statistical value indicating the likelihood of the vehicle type in the query image can be obtained.

b. **Intra-class Vehicle Classification (Sedan/Taxi classification)**

In this project, intra-class classification is carried out to classify two similar types of vehicle: Sedan car and Taxi. These two classes of vehicle are very much alike and therefore the classification system is modified from the inter-class classification system to fit the task.

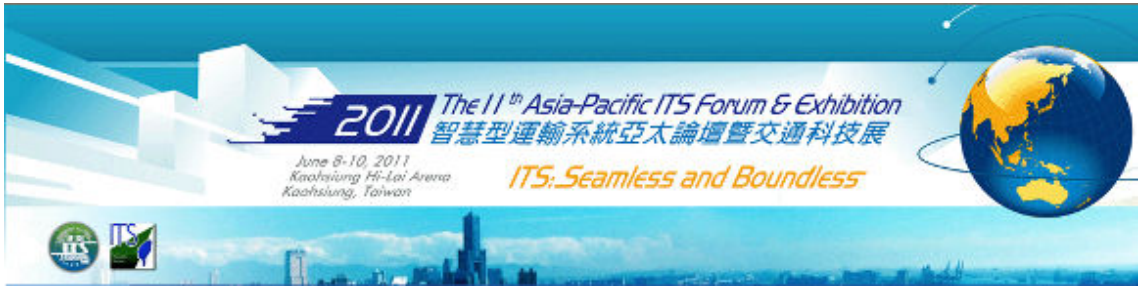

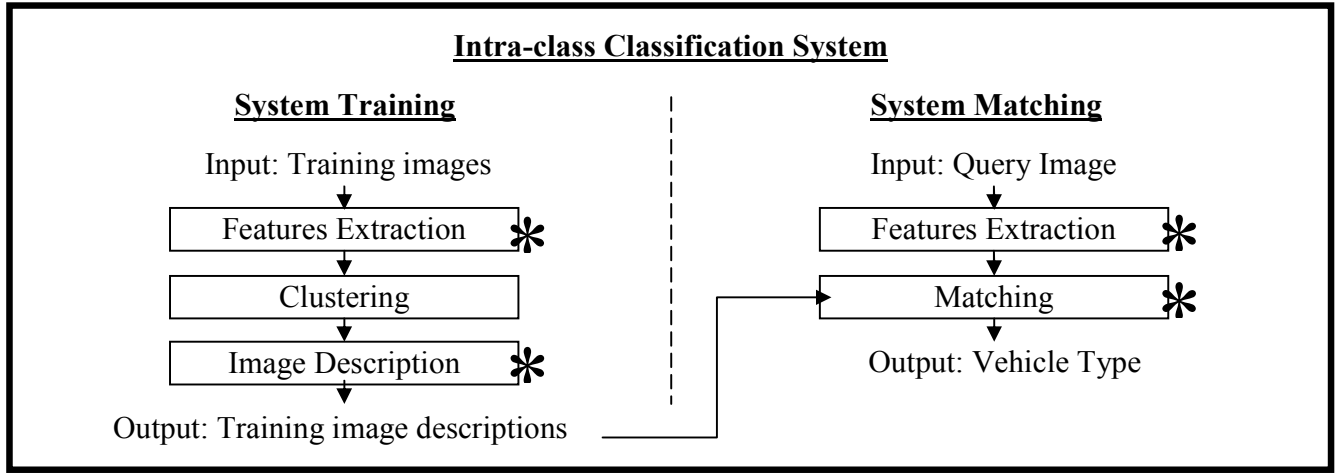

Figure 3.3: The training and matching scheme for the intra-class vehicle classification system

As shown in Figure 3.3 above, the training and matching scheme for the intra-class vehicle classification task has similar framework with the inter-class vehicle classification task despite some modification. These modifications are marked with asterisk in the Figure 3.3. The first modification took place in feature extraction stage. The modification is made to increase the number of features extraction to enhance the performance.

The major modification of the system takes place in the image description and matching stage. For inter-class vehicle classification, the features extracted from the training are assigned with weightage and used for matching. In intra-class vehicle classification, the features extracted from the training are used to describe the training images. Instead of using the clusters of features for the matching, the system creates the image description of each training images to serve as the matching medium.

In the matching scheme, the query image fed into the system undergoes features extraction. The features extracted are used to create the image description for the query image. The matching is carried out between the description of the query image and the descriptions of the training image. The vehicle type of the query image can be determined by the vehicle type of its matched training image.

## 4. Techniques

In this section, the techniques used in each components of the vehicle classification system are discussed thoroughly. These techniques include Modified SIFT, K-means clustering, Canny's Edge Detection and image description method which is proposed in this project.

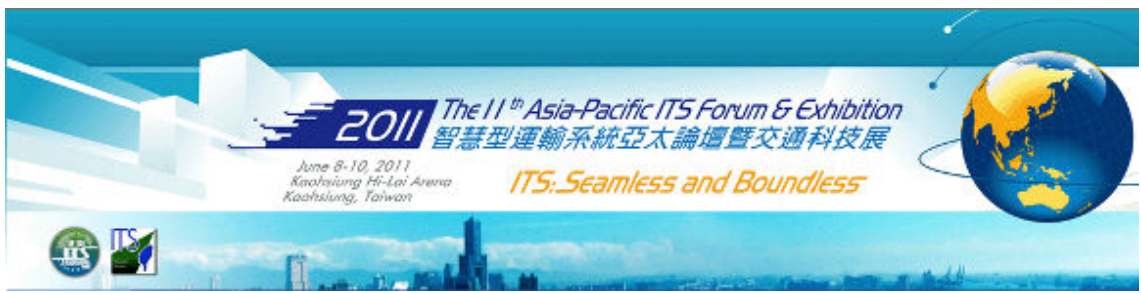

a. Features Extraction

In this component, the main tasks involves include the preprocessing of the image, detection of the keypoints, and the descriptions of the keypoints detected. For both two systems developed in this project, the preprocessing of the image involved are the same. The processing involved includes RGB to grayscale conversion and segmentation. RGB to grayscale conversion takes away the illumination details from the image to make the classification task independent of the illumination change. The segmentation task segment the non-vehicle region, leaving only the vehicle region on the image.

For inter-class classification task, the detection of keypoint is done using the edge line detection as proposed by Ma and Grimson [10]. In simpler words, the line detection method is employed to anchor the keypoints on the vehicle. The use of vehicle edge line as the keypoints anchor has proven to be repeatable and reliable. In this project, Canny's edge detection is used as the line detection algorithm. Every pixel that makes up the line at the end of the Canny's edge detection serves as the anchor of the keypoints.

For intra-class classification task, the keypoint detection is done without using the edge line detector. Instead, the entire pixels on the vehicle region after the segmentation serve as the anchor for the keypoints. This is because the low distinctiveness between the two classes of vehicle. More features are required for the classification task.

The feature detection methods as mentioned above only find out the position of the keypoint on the vehicle image. The keypoint have to be tagged with a value in order to be used for matching purpose. Here, the descriptor from SIFT is adopted to describe the keypoint. SIFT descriptor describe a keypoint based on the pattern around it [8, 9]. In this project, the SIFT descriptor is modified as proposed by Ma and Grimson [10]. The modification includes the use of unified scale and orientation for the description throughout the system. The modification may take away the robustness to orientation and scale, but the two advantages of SIFT is not of concern in such system. The descriptions of the keypoints serve as the matching medium for the system.

b. Clustering

In this project, K-means clustering is used to cluster the features extracted during the system training. The features extracted from the training images consist of all the features that can be found on the vehicles. These features can be redundant. Therefore, clustering method come into play to cluster these features into a specific number. For the systems in this project, four hundreds clusters are assigned for the clustering. Clustering not only helped the system to get rid of the redundancy of features, but also reduce the matching steps required for in the system matching.

c. Weightage Assignment

For inter-class vehicle classification task, all the feature clusters undergoes another matching with the training image. Unlike the matching in the matching scheme, this matching is done with the vehicle type of the training images known. This process is named as weightage assignment. As the name applies, this process assign a value which indicates the likelihood of the feature cluster belongs to a particular vehicle class type. This value is calculated using the formula shown in Figure 4.1. Figure 4.1 shows how a weightage value is calculated and assigned to a particular feature cluster. Each feature cluster are matched with *n* number of training images that belongs to a particular vehicle type. Then, the number of image matched, *m* is used to calculate the weightage value that indicates how likely the cluster is found on the particular vehicle type.

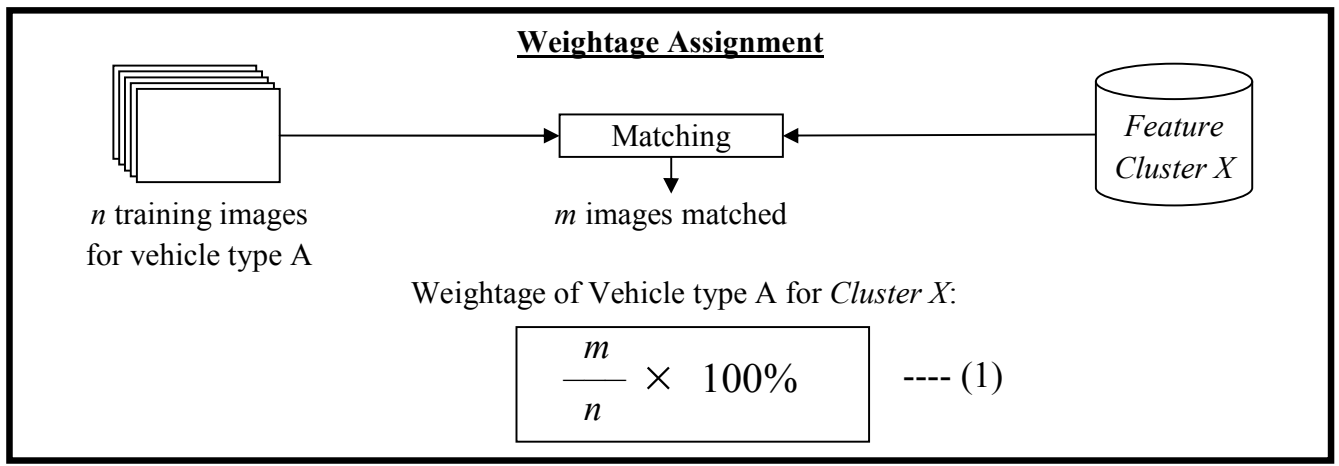

Figure 4.1: The weightage assignment and weightage calculation formula

d. **Image Description**

For intra-class vehicle classification, the matching medium is different. Instead of weighted clusters, each training images are described into an array of value indicating the existence of the clusters in the image. These image descriptions serve as the matching medium for the system.

As shown in Figure 4.2, each feature clusters are matched with the training image. The result of each match contributes to the value within an array. The length of the array equals to the number of the feature clusters. Each value indicates the existence of the feature within the image. When these values are arranged together into an array, the array can be highly distinctive. In this project, the number of feature clusters created is 400, which makes the total possible description as high as 2 to the power of 400. In other words, one description is has a probability of 1/ (2^400), which is generally almost zero and the description is very unique.

The concept of such description is similar to Human DNA. Human DNA sequences are so unique that a slightest match of a stretch can contribute to the matching of two human DNAs. In DNA analogy, the image description used in this project is the human DNA. A match of a query image description with the description of one of the training image would suggest the query image have the same vehicle type with the training image.

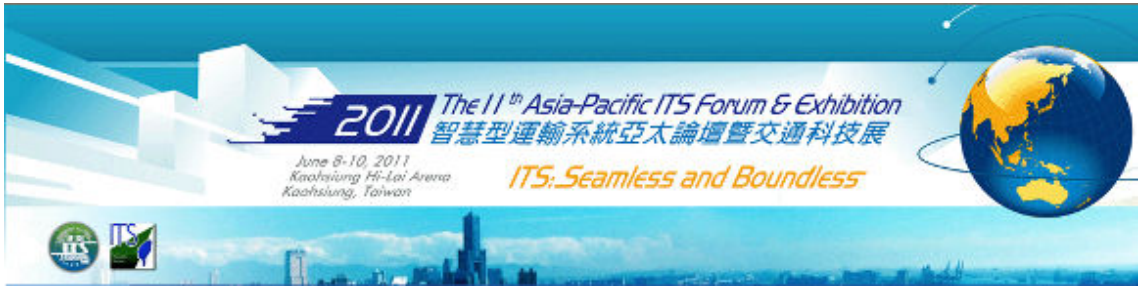

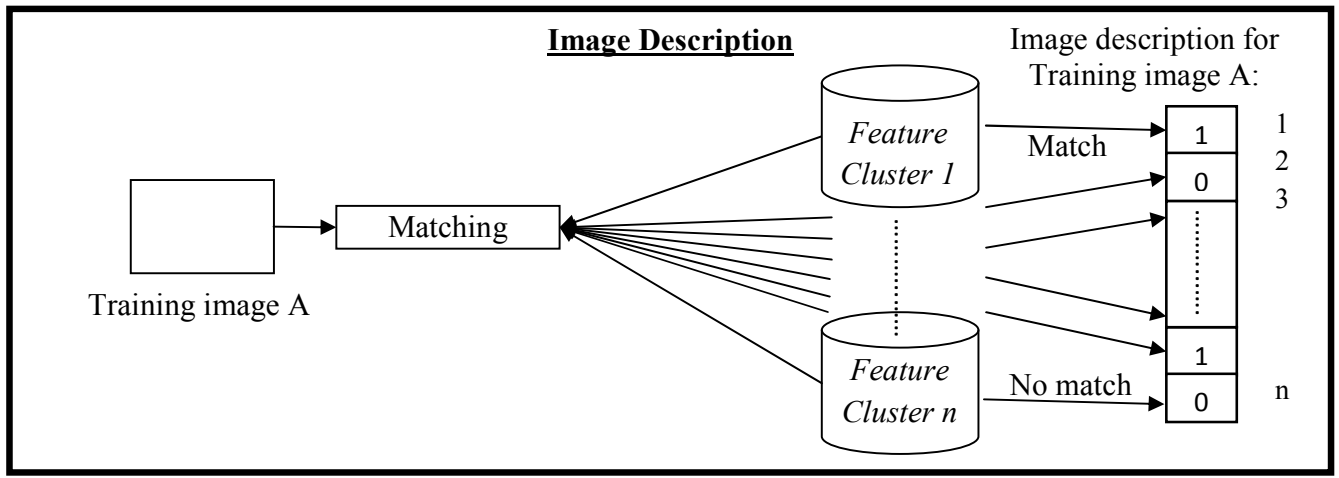

Figure 4.2: The image description

e.  **Matching**

The matching algorithm used in this project is the Euclidean distance matching, matching took place both in the training and matching scheme of the system. In system training, the matching took place in the weightage assignment and image description where the feature clusters are matched with the training image. Each query image will have two statistical values for two vehicle classes. When a feature cluster is matched on the query image, the weightage of the cluster towards each vehicle class is retrieved and place into account for the calculation of the statistical value. In system matching, the matching took place to match the query image with the output from the system training.

5.  **Result**

The system is tested with the open source dataset from Ma and Grimson [10] and returned promising result. The dataset is made up of 530 images of vehicle segmented from vehicle detection system, 200 images of sedan car, 200 images of van and 130 images of taxi. The images are captured from a mid-range surveillance camera and thus the lower quality and resolution.

For the training of system, 50 images of each class are randomly selected as the training image. Then, the whole dataset is feed into the system for classification. The result of the vehicle classification system obtained from the dataset is shown in the confusion matrix in table 5.1.

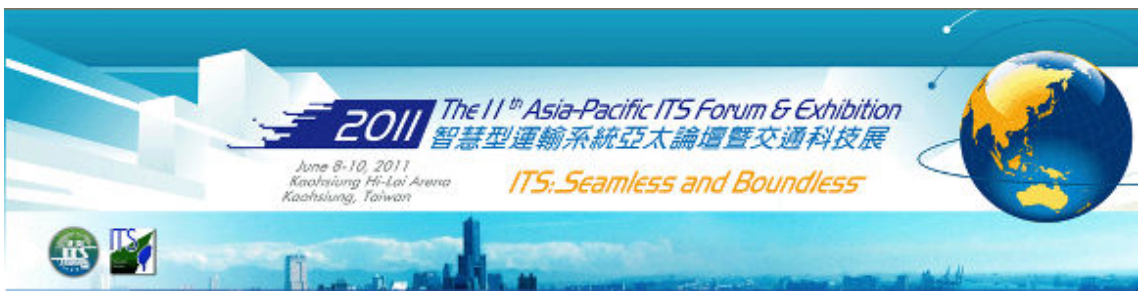

Table 5.1: Confusion matrix of the result obtained from the vehicle classification system

|  | Car | Minivan |
|---|---|---|
| Car | 98.5% | 1.5% |
| Minivan | 1.5% | 98.5% |

|  | Sedan | Taxi |
|---|---|---|
| Sedan | 93% | 7% |
| Taxi | 0.77% | 99.23% |

## 6. Conclusion

This project does not solely serves as the effort in finding a economic and efficient solution for the implementation of MLFF concept on road traffic, it has also broaden the room for research in related field. One conclusion be made was the potential of SIFT descriptor in image recognition. With suitable modification, the component from Lowe's SIFT has the potential to fit in many task in the vision-based task.

The weightage assignment and image description developed as the matching medium proposed in this project open up another room for improvement and research for image processing. The image description method which is analogous to human DNA concept is so unique that the probability of a description is so much near to zero when the array is long enough. This method can be used in area which involved low distinctive features such the intra-class vehicle classification in this project.

The system developed can also serve as the framework for any vehicle classification system. All in all, the project has been a success.

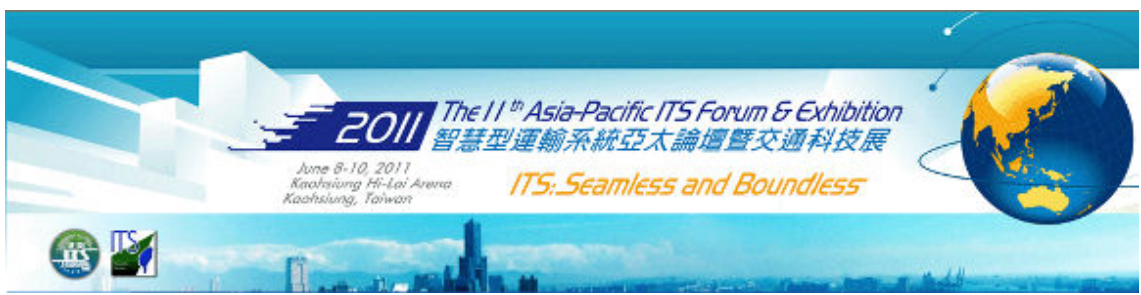